\documentclass[conference]{IEEEtran}
\usepackage{cite}
\usepackage{amsmath,amssymb,amsfonts}
\usepackage{algorithm} 
\usepackage{graphicx}
\usepackage{algpseudocode}
\usepackage{textcomp}
\usepackage{amsmath}
\usepackage{pdfpages}
\usepackage{booktabs}
\usepackage{graphicx}
\usepackage{float}          
\usepackage{caption}        
\usepackage{siunitx}
\usepackage{parskip}
\usepackage{graphicx}
\usepackage{subcaption}
\usepackage{multirow}
\usepackage{xcolor}
\def\BibTeX{{\rm B\kern-.05em{\sc i\kern-.025em b}\kern-.08em
    T\kern-.1667em\lower.7ex\hbox{E}\kern-.125emX}}
\begin{document}

\title{Neighbor-Aware View Synthesis for Restoring Missing Views in Light-Field Camera Arrays\\

}


\author{
\IEEEauthorblockN{Sakshi Goel, Ayush Goyal, K S Venkatesh, Koteswar Rao Jerripothula}
\IEEEauthorblockA{\textit{Indian Institute of Technology Kanpur (IIT Kanpur), India} \\
Email: \{sakshigl, gayush23, venkats, kotesrj\}@iitk.ac.in}
}

\maketitle

\begin{abstract}

In light-field (LF) imaging systems, dense spatial sampling from a camera array enables powerful post-capture capabilities such as refocusing and depth estimation. However, real-world LF capture is often affected by hardware malfunctions, where one or more cameras in the array fail, leading to missing sub-aperture images and degraded reconstruction quality. This paper addresses the problem of defective or missing view restoration in light-field camera arrays. We propose a novel generative framework that synthesizes the absent views by exploiting information from a carefully selected subset of neighboring cameras. These selected images, along with a positional encoding map indicating both their locations and the desired target view, are fed into a conditional Generative Adversarial Network (cGAN) trained to generate the missing viewpoint in a geometrically consistent manner. Extensive experiments on synthetic and real-world LF datasets demonstrate that our method produces visually plausible and photometrically accurate reconstructions, outperforming baselines for view interpolation both quantitatively and qualitatively. The proposed framework thus offers a robust and efficient solution for fault-tolerant light-field image acquisition.
\end{abstract}

\section{Introduction}
\label{sec:introduction}

Light field imaging has emerged as a revolutionary technology, attracting significant attention in both academia and industry, especially with the advent of commercial plenoptic cameras [1], [5] and recent dedication in the field of Virtual Reality (VR) and Augmented Reality (AR) [12]. Light field (LF) cameras need camera arrays, which can be incorporated in a single camera as well by placing a microlens array between the main lens and the image sensor, thereby capturing both the intensity and direction of light rays from real world scenes. 
While such camera array setups offer high spatial resolution and wide baselines, to achieve dense angular sampling using a synchronized grid of cameras, building a dense one is quite complex.

Moreover, the hardware malfunctions in the multi-camera architecture can lead to missing Sub-Aperture Images (SAIs), or simply views, leading to gaps in the 4D LF data structure. These missing views can significantly degrade performance of downstream algorithms resulting in incorrect depth maps and visible discontinuities in the final rendering.
One possible solution to this problem is missing view synthesis and is an important research direction to explore for building efficient and reliable LF acquisition systems.

Inspired by traditional approaches such as learnable interpolation or geometry based rendering, Kalantari et.al [4] breaks down the goal of view synthesis into disparity estimator and predictor modeled by Convolutional Nueral Network (CNN) and achieves it using the four corner views, outperforming prior methods [6], [7], but continues to struggle in reconstructing occluded and non-lambertain surfaces. Wu et.al [8] present a "blur-restoration deblur" framework using Epiplolar Plane Images (EPIs) without estimating the geometry of the scene. Recently, deep learning-based methods [10], [11], have shown remarkable success in image reconstruction and novel view synthesis. However, many existing frameworks do not promise efficient LF missing view restoration, as they may require large amounts of training data, may not explicitly leverage the underlying geometric consistency of the Light Field or exploit all the available input views.

\begin{figure}[t]
    \centering
    \includegraphics[width=0.9\columnwidth]{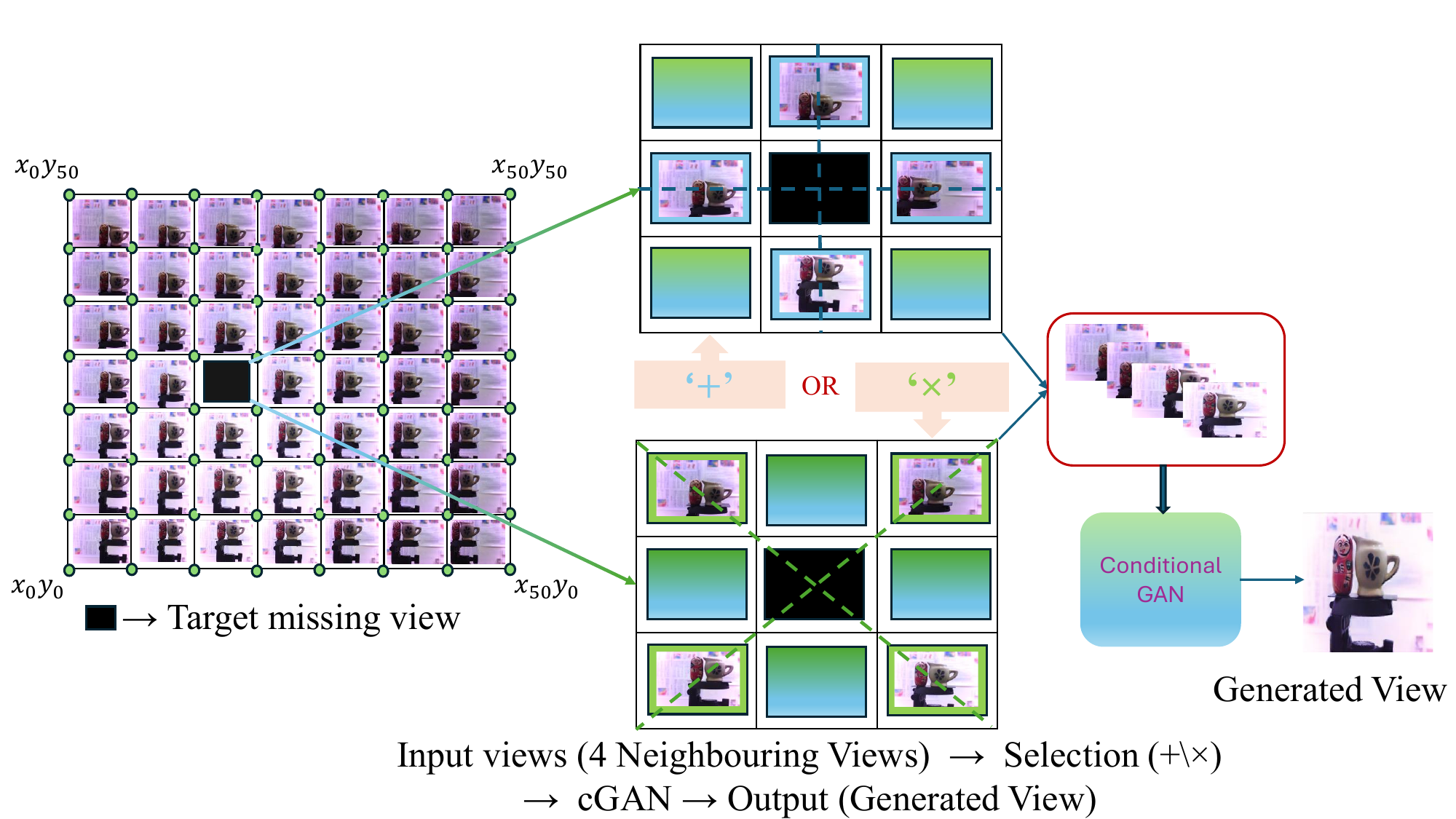}
    \caption{Light-field missing view synthesis using 4 neighboring views supplied to our novel conditional GAN architecture.}
    \label{fig:onecol}
\end{figure}

\begin{figure*}[t]
\centering
\includegraphics[width=\textwidth]{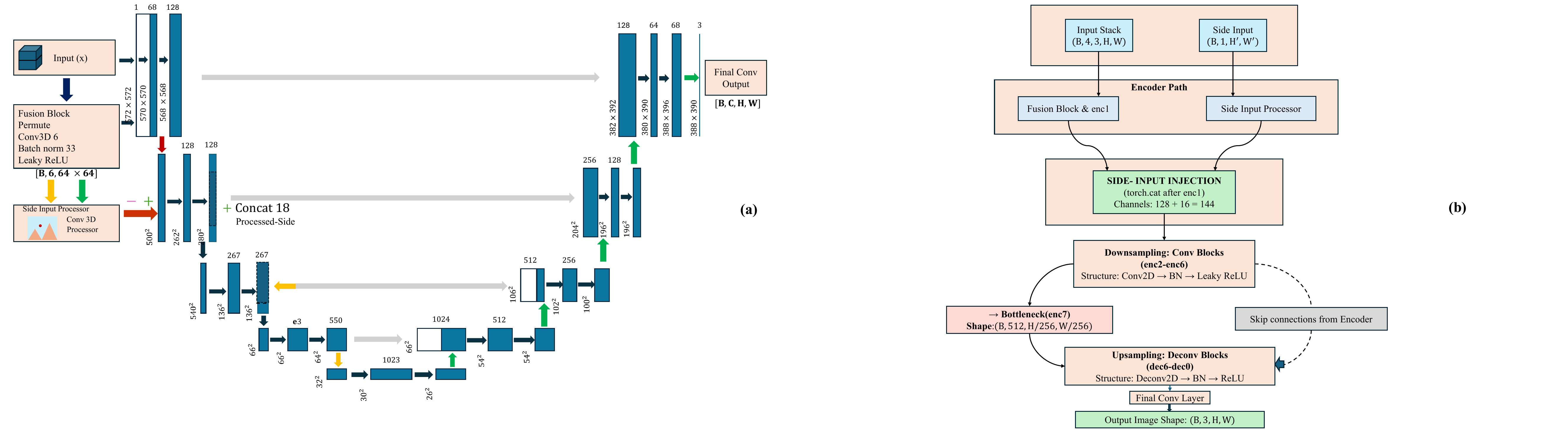}
\caption{Architecture of the proposed U-Net framework (a) U-Net Architecture (b) Model Structure}
\label{fig:unet_full}
\end{figure*}

To address the challenges of the LF missing view synthesis problem, this paper proposes a novel, neighbor-aware generative framework, as shown in Fig.~\ref{fig:onecol}. We argue that neighboring views capture relevant and accurate information, given that they share similar angular properties as the missed one. Moreover, our method is quite flexible/adaptive in choosing the 4 input views: any neighbor based on the availability along cardinal or diagonal axes within certain distance can be chosen. In contrast, [4] limits itself by having only corner views as input views. What if one of the corners is defective?

The main contributions of this work are listed below.
\begin{itemize}
    \item We propose a new synthetic dataset consisting of 121 Light Field scenes, covering a variety of objects, to serve as a challenging benchmark.  
    \item We propose a novel generative framework for light-field missing view restoration that relies on an adaptive selection of four neighboring views, making the system efficient and robust.
    \item We introduce a Positional Encoding Map as a key conditional input to our generator, providing an explicit geometric prior that significantly improves view synthesis consistency.
    \item We propose a novel 3D-2D fusion generator that effectively integrates multi-view information for high-fidelity image reconstruction.
\end{itemize}

\section{Methodology}
\label{sec:methodology}
\subsection{Approach Overview}
\label{subsec:Approach}

The approach proposed to restore the missing or defective views in Light Field data, is based on our novel conditional Generative Adversarial Network (cGAN), which carefully leverages the geometric and photometric redundancy present in LF data. The process can be decomposed as follows: an adaptive neighbor selection mechanism with positional encoding, a tailored conditional GAN architecture, and a robust loss function for exhaustive training and implementation.

\subsection{Adaptive Neighbor Selection and Positional Encoding}
\label{subsec:neighbor_selection}

Instead of indiscriminately using all the available input views from LF data, our framework initially starts with strategically selecting a geometrical optimal subset of input images. For a target missing view at angular coordinates $\mathbf{c}_{t} = (u_t, v_t)$, we adaptively identify four available neighboring SAIs and randomly choose one of the two configurations, `plus' ($\mathbf{+}$) or `x' ($\times$), and choose neighbors within a certain neighborhood. 

 \begin{figure}[!t]
     \centering
        \centering
        \includegraphics[width=0.9\linewidth]{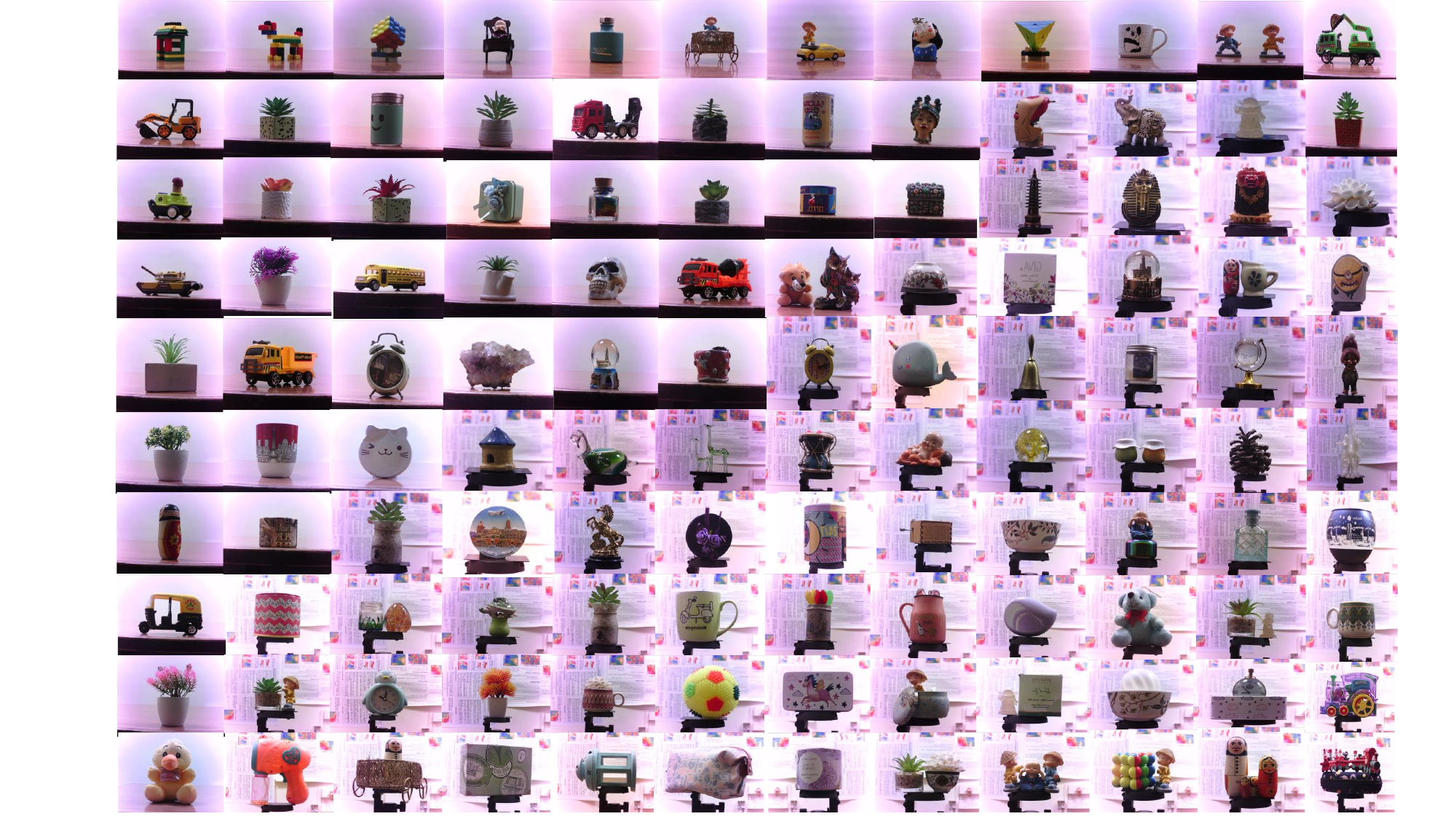}

     \caption{Central Views of scenes in our dataset}
     \label{fig:lf_fullpage_sidebyside}
\end{figure}


In Plus-Shape Configuration ($\mathbf{+}$), the four input neighbors are selected along the cardinal axes, one from each direction (Right, Left, Up, Down), each at a random distance between 1 to 10. Similarly, in X-Shape Configuration ($\times$), they are selected along the diagonal axes (Right-Up, Left-Up, Right-Down, Left-Down), each at a random distance between 1-10.

 The selected four neighboring SAIs, $\mathbf{I}_{i} \in \mathbb{R}^{3 \times H \times W}$ for $i \in \{1, 2, 3, 4\}$, are stacked along the channel dimension to form a 4D input tensor $\mathbf{X} \in \mathbb{R}^{4 \times 3 \times H \times W}$. 
We generate a \textbf{Positional Encoding map} as a Side Frame, $\mathbf{P}$, which is a low resolution single channel image that encodes the coordinates of the selected neighboring SAIs as per the configuration and the missing target view. 
We utilize the \textbf{Manhattan Distance} ($D_{manhattan}$) between the missing target view and its farthest neighbor to dynamically weight the reconstruction loss, enabling the network to focus and overcome the challenges in view synthesis.

\subsection{Conditional Generative Adversarial Network (cGAN)}
\label{subsec:cgan_architecture}
Our missing view restoration model is realized using a cGAN comprising of a specialized Generator $\mathcal{G}$ and a Patch-based Discriminator $\mathcal{D}$. This architecture is inspired by the Pix 2 Pix framework, conditioned on the input stack $\mathbf{X}$ of SAIs and the positional encoding map $\mathbf{P}$. Fig.~\ref{fig:unet_full} shows the architecture of the proposed U-Net framework.
\\

\subsubsection{Generator $\mathcal{G}$: UNet with 3D-2D Fusion}
\label{subsubsec:generator}
The Generator $\mathcal{G}$ follows a modified U-Net architecture (\texttt{UNetGenerator3DWithSideInput}) designed to effectively fuse multi-view information and geometric constraints:\begin{enumerate}
    \item \textbf{3D Fusion Block:} The input stack $\mathbf{X} \in \mathbb{R}^{B \times 4 \times 3 \times H \times W}$ is first permuted to $\mathbb{R}^{B \times 3 \times 4 \times H \times W}$ and processed by a 3D convolution layer with a kernel size of $(4, 4, 4)$ and a stride of $(1, 2, 2)$. This operation compresses the 4 input views into a single, fused feature map $\mathbf{F} \in \mathbb{R}^{B \times 64 \times H/2 \times W/2}$, effectively projecting the 4D LF data onto a 2D feature space while retaining multi-view spatial correlations. This fused map $\mathbf{F}$ serves as the initial feature representation ($e0$).
    \item \textbf{Side Input Processing:} The low-resolution Positional Encoding Map $\mathbf{P}$ ($\in \mathbb{R}^{B \times 1 \times 51 \times 51}$) is upsampled to the current feature map size ($64 \times 64$) and processed by a $2$-layer 2D convolutional block.
    \item \textbf{Adaptive Fusion in Encoder:} The processed side input is concatenated with the features after the first 2D downsampling layer. This adaptive fusion at an intermediate feature stage ensures the geometric prior guides the multi-view feature integration.
    \item \textbf{U-Net Structure:} The network then follows the standard U-Net encoding-decoding path with skip connections, ensuring detailed high-frequency information is preserved. The final output is the synthesized target view $\hat{\mathbf{I}}_{t} \in \mathbb{R}^{3 \times H \times W}$.
\end{enumerate}

\subsubsection{Discriminator $\mathcal{D}$: PatchGAN}
\label{subsubsec:discriminator}

We employ a \textbf{PatchGAN Discriminator} (\texttt{PatchGANDiscriminator}) to enforce local realism in the synthesized image. The discriminator takes the restored missing image $\hat{\mathbf{I}}_{t}$ (or the ground truth $\mathbf{I}_{t}$) and the conditional input, which is defined as the mean of the four neighboring SAIs ($\mathbf{X}_{avg} = \frac{1}{4} \sum \mathbf{I}_{i}$), concatenated along the channel dimension. The discriminator outputs a probability map which indicates the realness of local image patches, thereby driving the generator to produce photometrically and structurally plausible results.

\subsection{Training and Loss Functions}
\label{subsec:loss_function}

The cGAN is trained end-to-end to minimize a composite loss function $\mathcal{L}_{\mathcal{G}}$ for the Generator and $\mathcal{L}_{\mathcal{D}}$ for the Discriminator.

\subsubsection{Discriminator Loss}
The Discriminator loss $\mathcal{L}_{\mathcal{D}}$ is a standard adversarial loss based on Binary Cross-Entropy (BCE)[14]:
\begin{equation}
\label{eq:discriminator_loss}
\begin{split}
\mathcal{L}_{\mathcal{D}} = \frac{1}{2} \Biggl[ & \mathbb{E}_{\mathbf{I}_{t}} \left[ \log \mathcal{D}(\mathbf{X}_{avg}, \mathbf{I}_{t}) \right] \\
 & + \mathbb{E}_{\mathbf{X}, \mathbf{P}} \left[ \log \left( 1 - \mathcal{D}(\mathbf{X}_{avg}, \mathcal{G}(\mathbf{X}, \mathbf{P})) \right) \right] \Biggr]
\end{split}
\end{equation}

\subsubsection{Generator Loss}
The Generator loss $\mathcal{L}_{\mathcal{G}}$ is a weighted combination of adversarial, reconstruction, structural, and perceptual terms [9][14][15][16]:
\begin{equation}
\begin{split}
\mathcal{L}_{\mathcal{G}} 
= \mathcal{L}_{GAN} + \lambda_{L1} \mathcal{L}_{L1}
+ \lambda_{SSIM} \mathcal{L}_{SSIM} \\
+ \lambda_{Perc} \mathcal{L}_{Perc} + \lambda_{Edge} \mathcal{L}_{Edge}
\end{split}
\end{equation}

\textbf{1. Adversarial Loss ($\mathcal{L}_{GAN}$):}
This term encourages the synthesized image $\hat{\mathbf{I}}_{t}$ to fool the Discriminator:
\begin{equation}
\mathcal{L}_{GAN} = \mathbb{E}_{\mathbf{X}, \mathbf{P}} \left[ \log \left( 1 - \mathcal{D}(\mathbf{X}_{avg}, \mathcal{G}(\mathbf{X}, \mathbf{P})) \right) \right]
\end{equation}

\textbf{2. Weighted Reconstruction Loss ($\mathcal{L}_{L1}$ and $\mathcal{L}_{SSIM}$):}
To enforce photometric accuracy, we use a combination of $L1$ loss and Structural Similarity Index Measure (SSIM) loss, both weighted by the inverse difficulty of the synthesis task, represented by the Manhattan distance of the target view:
\begin{equation}
\mathbf{W} = 1.0 + \lambda_{dist} \cdot \frac{D_{manhattan}}{D_{max}}
\end{equation}
\begin{itemize}
    \item \textbf{Weighted L1 Loss:}
    $$
    \mathcal{L}_{L1} = \mathbb{E} \left[ \mathbf{W} \cdot \left\| \hat{\mathbf{I}}_{t} - \mathbf{I}_{t} \right\|_{1} \right]
    $$
    \item \textbf{Weighted SSIM Loss:} SSIM is calculated on images normalized to $[0, 1]$.
    $$
    \mathcal{L}_{SSIM} = \mathbb{E} \left[ \mathbf{W} \cdot \left( 1 - \text{SSIM}(\hat{\mathbf{I}}_{t}, \mathbf{I}_{t}) \right) \right]
    $$
\end{itemize}
\begin{figure}[!t]
    \centering

        \includegraphics[width=0.9\linewidth]{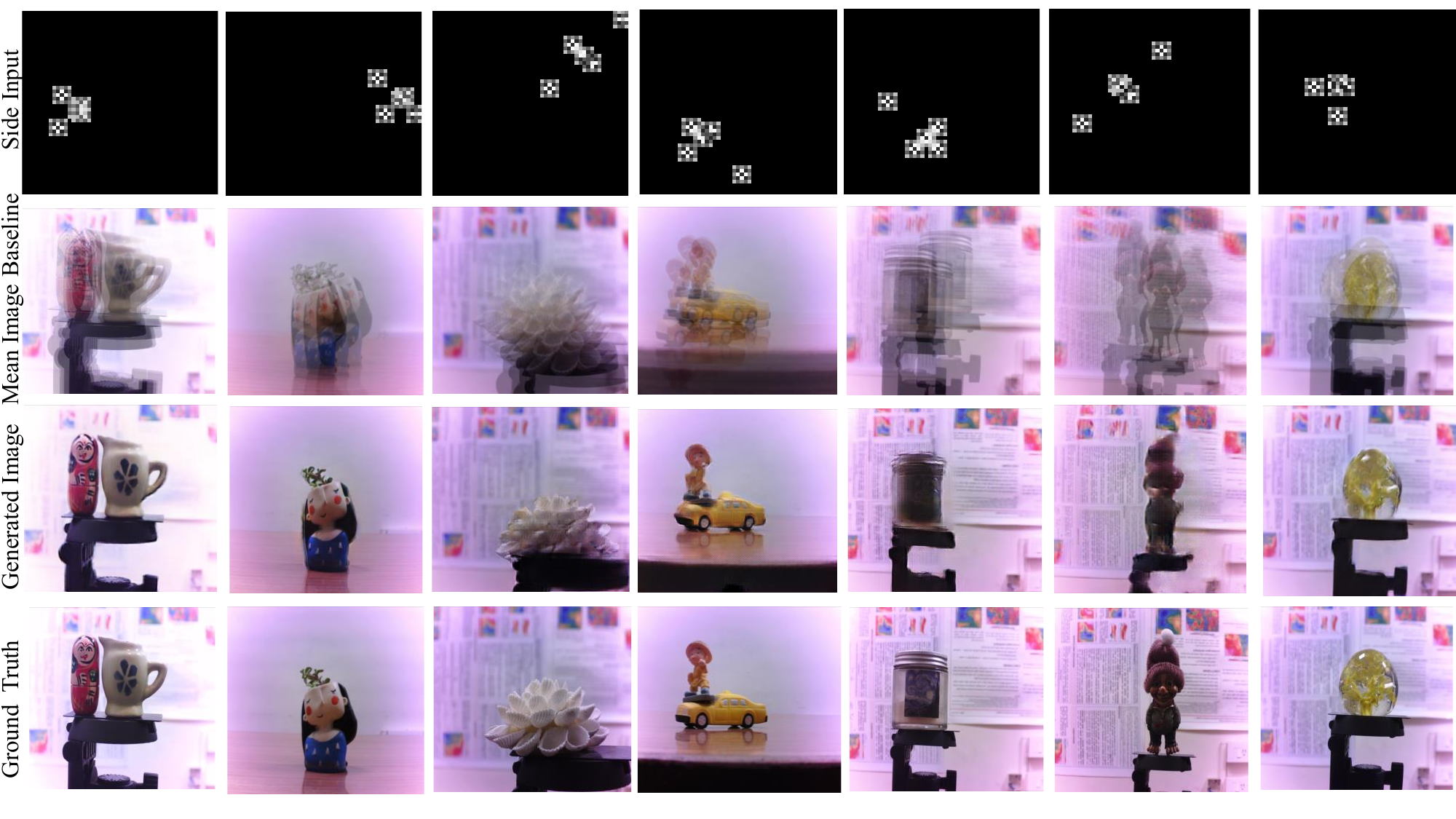}


    \caption{Visual Comparison for different Light Field Scenes}
    \label{fig:lf_full1}
\end{figure}



\textbf{3. Perceptual Loss ($\mathcal{L}_{Perc}$):}
To improve perceptual quality and visual plausibility, we introduce a VGG-19 based perceptual loss. This loss minimizes the $L1$ distance between the feature representations of the generated and ground truth images extracted from specific layers of a pre-trained VGG-19 network:
\begin{equation}
\mathcal{L}_{Perc} = \sum_{j} \left\| \phi_{j}(\hat{\mathbf{I}}_{t}) - \phi_{j}(\mathbf{I}_{t}) \right\|_{1}
\end{equation}
where $\phi_{j}(\cdot)$ denotes the feature map extracted from the $j$-th selected layer of the VGG network.
\\
\\
\textbf{4. Edge Loss ($\mathcal{L}_{Edge}$)}:
During model evaluation, a recurring vertical blur artifact was observed in the generated images. To mitigate this and enforce sharper, more defined object boundaries, a dedicated Edge Loss was introduced. We use a pre-trained model, TEED, [9] as an expert edge extractor, to compute edge maps for both the generated image ($\hat{\mathbf{I}}_t$) and the ground truth target ($\mathbf{I}_t$). The loss function adapts TEED's proposed ``Double Loss'' ($\mathcal{L}_{dloss}$), which combines two distinct components to ensure both the completeness and quality of the detected edges. A Weighted Cross-Entropy loss ($\mathcal{L}_{wce}$) is applied to intermediate edge maps to encourage the detection of all possible edges, while a Tracing Loss ($\mathcal{L}_{trcg}$) is applied to the final fused edge map to promote thinner and clearer boundaries. By minimizing the difference between the edge map hierarchies of the generated and target images, this loss provides a strong, explicit gradient that guides the generator to produce structurally coherent images with crisp outlines.

The key loss weights used in the training are: $\lambda_{L1}=10.0$, $\lambda_{SSIM}=5.0$, $\lambda_{Perc}=1.0$, and $\lambda_{dist}=1.0$.

\section{Experiments and Results}
\label{sec:experiments}
\subsection{Datasets and Implementaion Details}
\label{subsec:datasets}
\textbf{Datasets}: We capture our light-field scenes using a cost-effective system using a precision ball screw–based 2D actuation platform with a single moving camera. Designed for static scenes, the setup captures high-resolution images (51×51 views) with 2 mm horizontal and 4 mm vertical steps, controlled via Arduino and MATLAB for automated movement and sequential capture. Built using three industrial axes supporting up to 1 kg payload, the platform ensures high positioning accuracy. Calibration is performed by selecting four extreme corners of the scene to define the capture area, with horizontal and vertical steps computed for uniform coverage  for better light field reconstruction and view synthesis. Fig.~\ref{fig:lf_fullpage_sidebyside} provides Central Views of captured 121 Light Field Scenes. 
We also use MPI Light Field Intrinsic Dataset [13], containing complex scenes with varying depths and lighting.

All models were trained on an NVIDIA RTX A6000 GPU for 200 epochs using the Adam optimizer with a learning rate of \(2\times10^{-4}\) and betas \((\beta_1,\beta_2)=(0.5,0.999)\). We used an 80/20 train-test split.

\begin{table}[t]
\centering
\caption{Quantitative comparison of on our and MPI datasets. The proposed model consistently achieves the best performance across all metrics.}
\setlength{\tabcolsep}{3pt} 
\renewcommand{\arraystretch}{1.1} 
\begin{tabular}{l@{\hskip 3pt}ccc@{\hskip 6pt}ccc}
\toprule
\multirow{2}{*}{\textbf{Model}} & \multicolumn{3}{c}{\textbf{Our Dataset}} & \multicolumn{3}{c}{\textbf{MPI Dataset}} \\
\cmidrule(lr){2-4} \cmidrule(lr){5-7}
 & \scriptsize SSIM$\uparrow$ & \scriptsize PSNR$\uparrow$ & \scriptsize LPIPS$\downarrow$ & \scriptsize SSIM$\uparrow$ & \scriptsize PSNR$\uparrow$ & \scriptsize LPIPS$\downarrow$ \\
\midrule
Baseline (Avg 4 Inputs) & 0.6519 & 19.58 & 0.2582 & 0.6337 & 21.12 & 0.2778 \\
3D U-Net & 0.6792 & 13.73 & 0.2421 & 0.6232 & 17.73 & 0.2245 \\
3D PatchGAN & 0.6833 & 23.61 & 0.3298 & 0.6367 & 20.72 & 0.3472 \\
3D PatchGAN + Side-Input & 0.7244 & 22.43 & 0.1430 & 0.6712 & 21.34 & 0.1323 \\
\textbf{Proposed} & \textbf{0.7572} & \textbf{24.24} & \textbf{0.1023} & \textbf{0.7006} & \textbf{22.51} & \textbf{0.1184} \\
\bottomrule
\end{tabular}
\label{tab:combined_results}
\end{table}

\subsection{Evaluation Metrics and Baselines}
\label{subsec:metrics}

We employ three standard metrics to quantitatively evaluate the quality of the synthesized views against the ground truth images: Peak Signal-to-Noise Ratio (PSNR), Structural Similarity Index Measure (SSIM), and Learned Perceptual Image Patch Similarity (LPIPS).   

As a simple baseline for comparison, we compute the pixel-wise average of the four selected neighboring input Sub-Aperture Images. This baseline helps to quantify the improvement gained from our learning-based approach over trivial interpolation.

\subsection{Ablation Study and Quantitative Analysis}
\label{subsec:ablation_study}

To analyze the contribution of each key component in our framework, we performed an ablation study by systematically building up our final model. We started with a basic U-Net generator and progressively added the discriminator, the positional encoding side-input, and the full composite loss function. \textbf{3D U-Net:} We first trained our generator architecture as a standalone model, using only the weighted \(L_{1}\) reconstruction loss. This configuration serves to evaluate the raw synthesis capability of the 3D-2D fusion U-Net. \textbf{3D PatchGAN:} Next, we incorporated the PatchGAN discriminator and trained the model within the complete cGAN framework using an adversarial loss combined with the \(L_{1}\) loss. This step assesses the improvement in image realism and sharpness from adversarial training. \textbf{3D PatchGAN with Side-Input:} We then integrated the Positional Encoding Map (P) as a conditional side-input to the generator. This variant tests our core hypothesis that providing explicit geometric guidance is crucial for accurate view synthesis. \textbf{Proposed (Full Model):} Finally, we trained our complete model as described in Section~\ref{subsec:loss_function}, utilizing the full composite loss function which includes adversarial, weighted reconstruction (\(L_{1}+L_{\mathrm{SSIM}} + L_{\mathrm{edge}}\)), and perceptual (\(L_{\mathrm{Perc}}\)) terms. This represents our final, optimized framework.

\subsection{Discussion}
\label{subsec:discussion}
The qualitative comparisons in Figs.~\ref{fig:lf_full1} demonstrate the effectiveness of the proposed model in reconstructing missing light-field views. Our generated results preserve fine spatial details, exhibit sharper object boundaries, and maintain consistent illumination across views. The inclusion of positional encoding clearly enhances geometric accuracy, aligning the synthesized structures more faithfully with the ground truth. These improvements highlight the model’s ability to leverage multi-view and geometric cues for photometrically and structurally coherent view synthesis.\\
The quantitative results of our ablation study are summarized in Table \ref{tab:combined_results}. The results clearly demonstrate the incremental benefits of each proposed component. The baseline averaging method yields poor results, highlighting the complexity of the task. While a basic 3D U-Net provides a reasonable starting point, incorporating the PatchGAN discriminator significantly boosts PSNR. The most substantial leap in performance, especially in SSIM and LPIPS, is observed upon introducing the positional encoding side-input, confirming its critical role in preserving geometric consistency. Our final model, trained with the comprehensive loss function, achieves the best scores across all metrics, producing images that are not only pixel-wise accurate but also structurally coherent and perceptually convincing.

\section{Conclusion}
This paper addresses the challenge of restoring missing views in light-field camera arrays due to hardware failures. It introduces a novel conditional GAN framework that uses a Positional Encoding Map and a 3D-2D fusion generator to synthesize accurate and geometrically consistent views from sparse neighboring inputs. Extensive experiments show impressive results.  


\end{document}